\documentclass{article}

\usepackage{arxiv}

\usepackage[utf8]{inputenc} 
\usepackage[T1]{fontenc}    
\usepackage[pagebackref=true,breaklinks=true,letterpaper=true,colorlinks,bookmarks=false]{hyperref}
\usepackage{url}            
\usepackage{booktabs}       
\usepackage{amsfonts}       
\usepackage{nicefrac}       
\usepackage{microtype}      
\usepackage{lipsum}
\usepackage{graphicx}
\usepackage{multirow}
\usepackage{amsmath}
\usepackage{amssymb}
\usepackage{bbm}
\usepackage{subcaption}
\graphicspath{ {./images/} }

\title{Semi-Supervised Few-Shot Atomic Action Recognition}

\author{
 Xiaoyuan Ni \thanks{Equal contribution.} \\
 Hong Kong University of Science and Technology\\
  \texttt{xniac@connect.ust.hk} \\
   \And
 Sizhe Song \footnotemark[1] \\
 Hong Kong University of Science and Technology\\
   \texttt{ssongad@connect.ust.hk} \\
  \And
 Yu-Wing Tai \\
 Kauishou\\
  \texttt{yuwing@gmail.com} \\
   \AND
 Chi-Keung Tang\\
 Hong Kong University of Science and Technology\\
   \texttt{cktang@cs.ust.hk} \\
}

\begin{document}
\maketitle
\bibliographystyle{unsrt}
\begin{abstract}
Despite excellent progress has been made, the performance on action recognition still heavily relies on specific datasets, which are difficult to extend new action classes due to labor-intensive labeling. Moreover, the high diversity in Spatio-temporal appearance requires robust and representative action feature aggregation and attention.  To address the above issues, we focus on atomic actions and propose a novel model for semi-supervised few-shot atomic action recognition.  Our model features unsupervised and contrastive video embedding, loose action alignment, multi-head feature comparison, and attention-based aggregation, together of which enables action recognition with only a few training examples through extracting more representative features and allowing flexibility in spatial and temporal alignment and variations in the action.  Experiments show that our model can attain high accuracy on representative atomic action datasets outperforming their respective state-of-the-art classification accuracy in full supervision setting.
\end{abstract}


\section{Introduction}
Significant achievements have been made in  action recognition~\cite{video-classification-cnn,action-forecasting,ucf} thanks to the development of 3D-convolutional networks (C3D) and Spatio-temporal attention over videos. 
Recent sequential embedding networks including LSTM~\cite{lstm} and temporal convolution (TCN)~\cite{tcn} have been applied for achieving better temporal alignment of video action.
However, the performance of state-of-the-art action recognition models relies on large-scale training datasets which are not easy to collect and annotate. 
In particular, the pertinent action may not occupy the entire Spatio-temporal volume of the given video, i.e., the action may occupy a subset of spatial and temporal volume of the given video frames with intra-class variations in relative position and length. 
To further complicate the problem, the relative sequences of sub-actions may vary, i.e., ‘playing basketball’ may contain a different permutation of ‘dribbling’ and ‘passing’, which poses great challenges in temporal alignment. 

Current methods either ignore alignment such as permutation-invariant attention~\cite{arn} or impose overly strict alignment such as dynamic time warp~\cite{otam}. The flexibility within action also presents difficulty in the aggregation of action features within a class. Naïve aggregation functions such as summation may harm representation which may also be easily affected by outliers.

To tackle the above issues, this paper focuses on {\em atomic} or fine-grained actions of duration typically less than 2 secs (e.g., dribbling and passing), which sidestep the need for strict alignment while making loose alignment sufficient. Atomic actions have shown promising performance gain for action recognition over conventional methods trained using coarse-grained action videos (e.g., playing basketball)~\cite{haa}. 
We propose a novel semi-supervised network for the few-shot atomic action classification, that supports action recognition of long query videos under the $K$-way $N$-shot setting~\cite{meta-learning}. Specifically, our model features a better understanding of human-centric atomic action with:
\begin{enumerate}
\item \textit{Semi-supervised training.} The video embedding module is trained in an unsupervised manner, extracting representative video features and classifying the action given only several examples.
\item \textit{Loose action alignment.} We adopt sliding windows over the temporal domain and use connectionist temporal classification (CTC) loss~\cite{ctc} to train the video embedding with relatively loose alignment, making the model more robust to variations in the sub-action permutation.
\item \textit{Multi-head video comparison.} We develop a multi-head relational network to consider
both global and local similarity.
\item \textit{Attention-based feature aggregation.} Our model aggregates class features through computing the mutual relationship between support and query videos. Compared with traditional mean aggregation, the attention-based one extracts the most related features for classification, thus reducing the representation complexity for class features and improving classification efficiency. \end{enumerate}

Overall, this paper contributes to 
few-shot atomic action recognition with semi-supervised learning.
Extensive experiments over published datasets show that our method outperforms the state-of-the-art accuracy achieved by models trained in full supervision.

\begin{figure*}[t]
    \begin{center}
        \includegraphics[width=\textwidth]{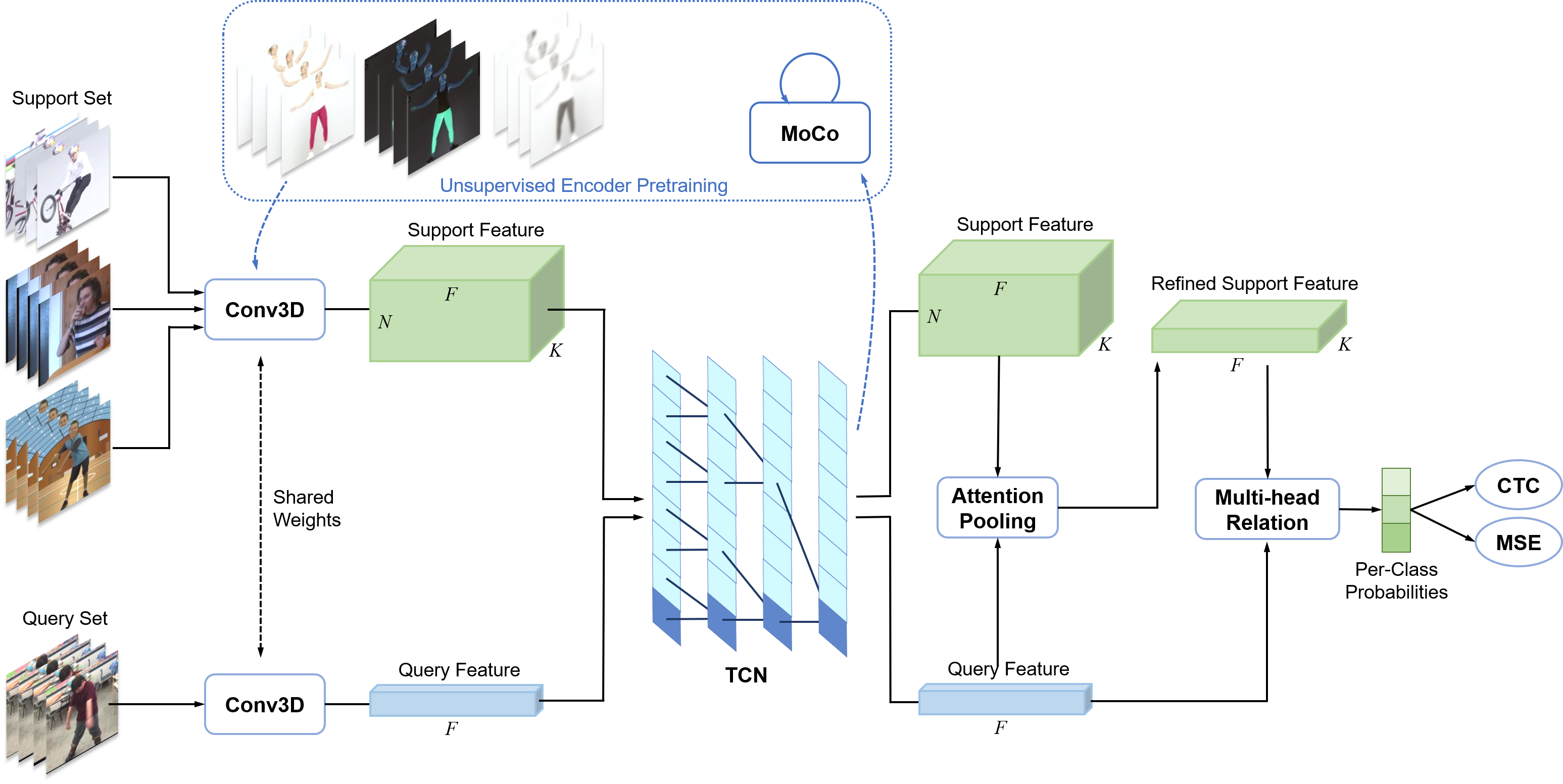}
    \end{center}
    \caption{\textbf{Overview.} Our model provides fine-grained spatial and temporal video processing with high length flexibility, which embeds the video feature and temporally combines the features with TCN. Further, our models provides  attention pooling and compares the multi-head relation. Finally, the CTC and MSE loss enables our model for time-invariant few shot classification training.}
    \label{fig:architecture}
\end{figure*}

\section{Related Work}
\subsection{Few Shot Learning}
Few Shot Learning (FSL) refers to a machine learning strategy that learns from a small amount of labeled data where the data labeling cost on large datasets can be prohibitively high~\cite{fsl}. Wide use of FSL includes multiple object tracking and detection~\cite{fs-mots,fs-od} and gesture recognition~\cite{boltzmann,sign-language}. 

In this paper, we propose FSL with novel technical contributions in embedding complex human actions through extending the original relational network~\cite{relation-net} into a multi-head relation network for robust feature comparison that adequately takes into consideration the high variety of atomic actions, while not requiring a large amount of human-annotated data. 


\subsection{Action Recognition}
\label{action-recognition}
Significant progress has been made in action recognition and classification with the recent development of 3D convolutional models, e.g., I3D~\cite{kinetics}, C3D~\cite{c3d} and TRN~\cite{trn}. All these models perform data-driven classification and process fixed-size videos by combining the locality in either the temporal or spatial domain. Their high accuracy is highly dependent on dependency on datasets used in training and testing.

To address the data issue, we experimented with a number of FSL for action recognition and found that almost all of these works attempt to align the video in the temporal domain and matching the relative 2D frames rather than 3D videos~\cite{taen,otam}, or search for the temporal attention of video~\cite{arn}. While achieving temporal alignment and attention techniques, these methods partition the given video into individual frames or tiny clips, thus introducing great complexity in their alignment strategies and inevitably losing the generality over datasets with distinct Spatio-temporal features. In contrast, our method provides a simple model with holistic understanding over the entire video, 
focusing on the human-centric video prediction without relying on any background and object information.

\subsection{Semi-Supervised Learning}
\label{moco_intro}
Semi-supervised learning is the learning based on both labeled and unlabeled data. In our task, although all the videos all have action labels, there are no boxes to localize where the actions are taking place in individual frames.  Thus it is possible to divide the learning strategies into two stages: the first stage is action classification with supervised learning and the second stage is action localization in terms of spatial attention with unsupervised learning. In~\cite{arn}, spatial attention, and temporal attention are trained with unsupervised learning.

Typical issues in applying unsupervised learning in feature extraction include limited dictionary size and inconsistent memory. Most recently, the Momentum Contrast (MoCo) has been proposed for unsupervised visual representation learning~\cite{moco}, which regards contrastive learning as dictionary-lookup and builds a dynamic and consistent dictionary on-the-fly. In this paper, MoCo has adopted to pretrain our encoder under an unsupervised setting.

\section{Method}
Figure~\ref{fig:architecture} illustrates our model structure.
We use a C3D+TCN encoder to embed a video clip to obtain the input feature. The C3D extracts Spatio-temporal features from videos and TCN processes the temporal information on a larger scale. Next, we apply an attention-pooling module where the support features are refined and integrated. With the query features and refined support features of each class, we then compute the classification probability by a multi-head relation network. Finally, the probability vector and ground truth label are used to obtain a CTC loss and MSE loss to update the network.

\subsection{Action Augmentation}
\label{augmentation}
We apply the following three augmentation methods:
\begin{enumerate}
    \item {\em Human-centric Cropping and Normalization}. Process the videos to produce human-centric video tubes and normalize the figures to the center, expecting the encoder can thus learn to lay more emphasis on human bodies in the videos, and invariant to background information.
    \item {\em Background Subtraction}. Apply background subtraction where the moving object is regarded as foreground and the rest is simply discarded. 
    \item {\em Usual Image Augmentation}. Apply random flipping, blurring, color inverting and rotation over each frame. 
\end{enumerate}

\begin{figure}[h]
        \includegraphics[width=0.49\textwidth]{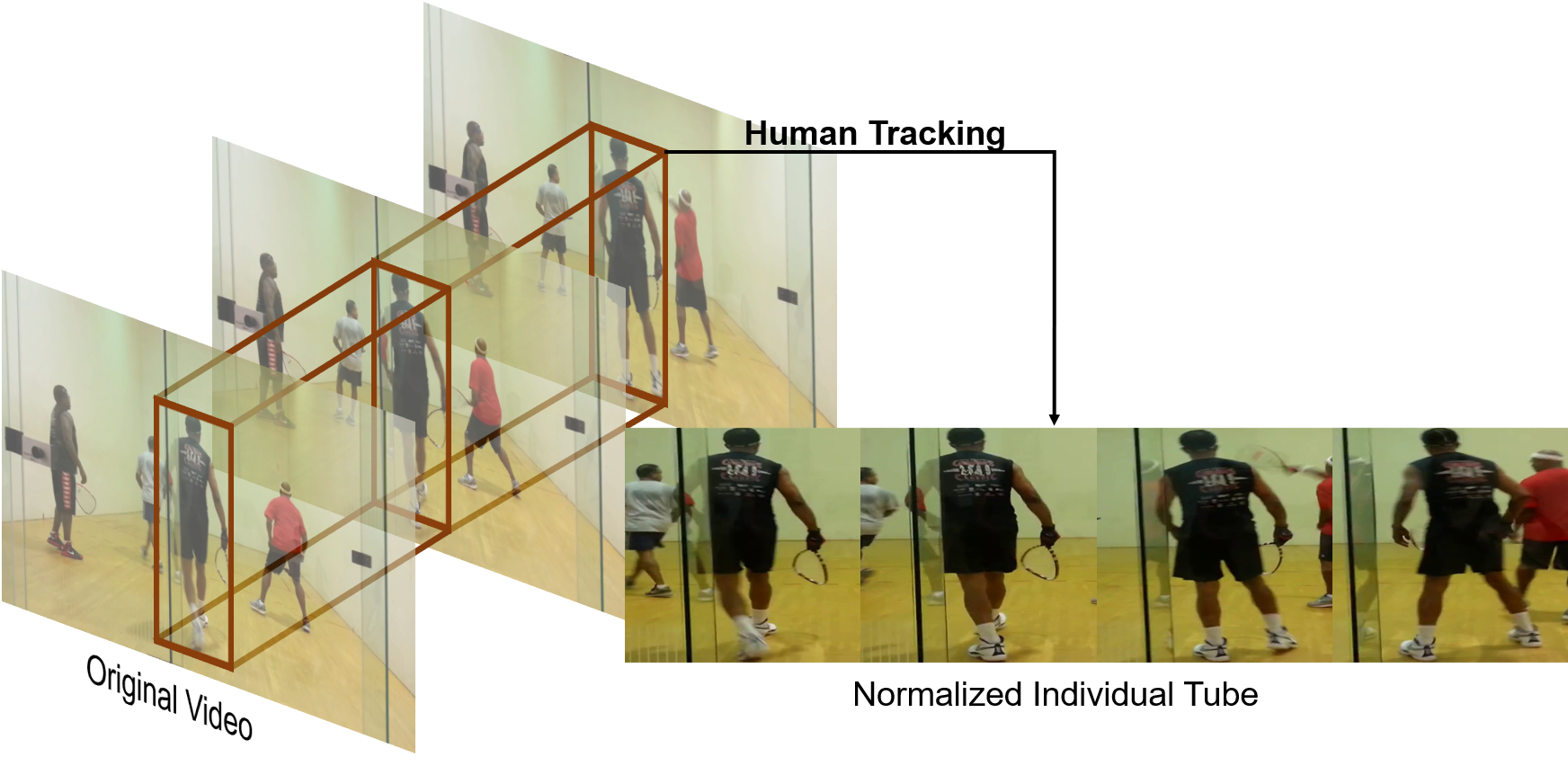}
        \includegraphics[width=0.18\textwidth]{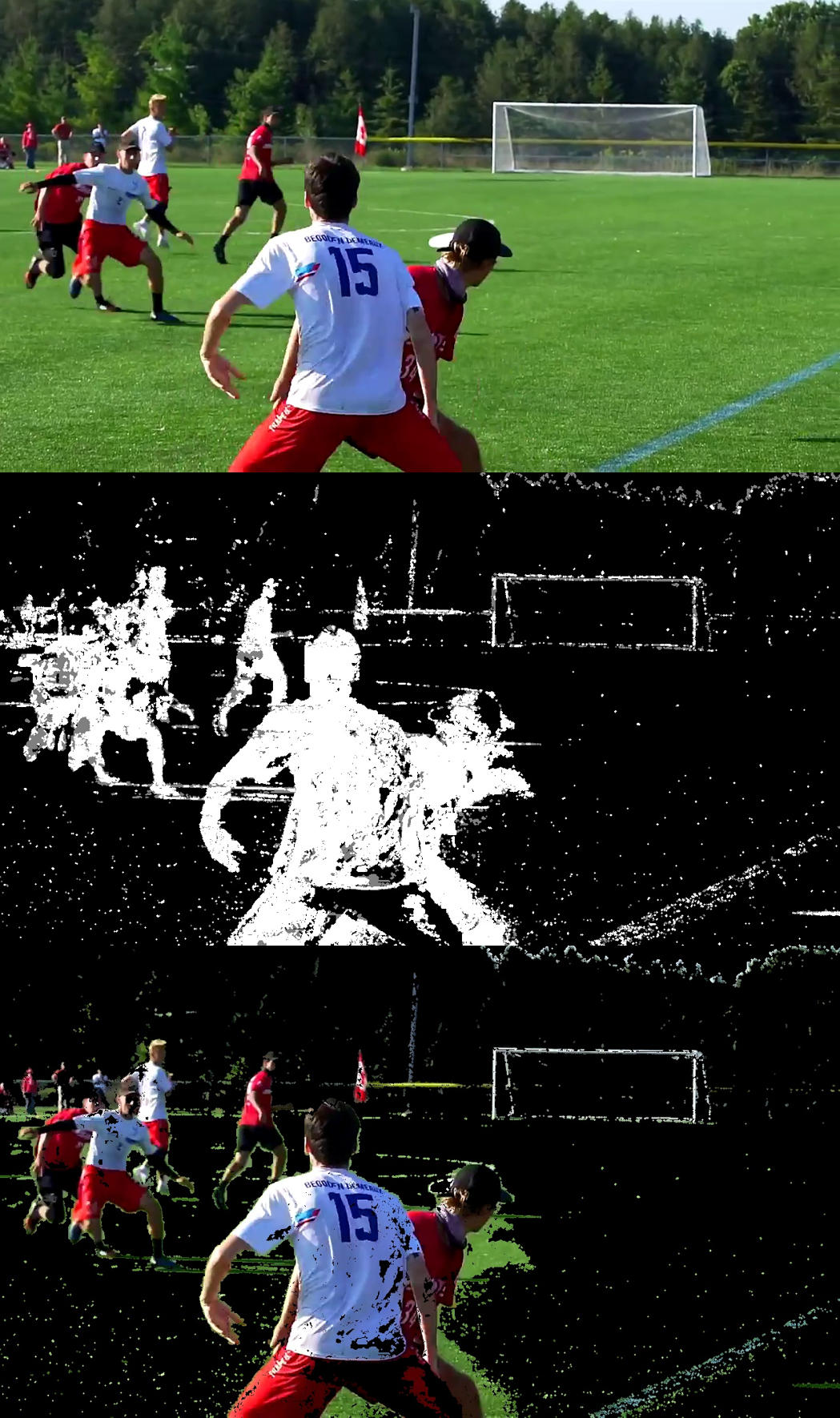}
        \includegraphics[width=0.32\textwidth]{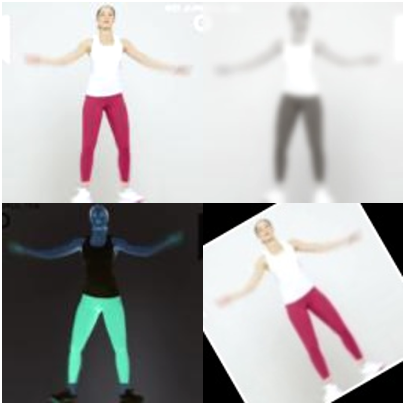}
        \begin{minipage}[t]{0.45\linewidth}\centering\subcaption{Human-centric}\label{fig:mot}\end{minipage}
        \begin{minipage}[t]{0.23\linewidth}\centering\subcaption{Background Subtraction}\end{minipage}
        \begin{minipage}[t]{0.32\linewidth}\centering\subcaption{Usual Image Augmentation}\end{minipage}
    \vspace{-7pt}
    \caption{Three methods of action augmentation.}
\end{figure}
The last two methods are easy to implement. For the first method, see Figure~\ref{fig:mot} where we use a human tracking tool, such as real-time MOT~\cite{realtimeMOT} to obtain human-centric video tubes, which are frame-wise bounding boxes of the human body with individual's identity labeled in each box. Given these video human tubes, we crop the original frames to get normalized images that precisely capture the movement of each individual in the video. While real-time MOT can generate all the normalized frames on-the-fly, in our modular implementation we generate all the tubes in advance.

\subsection{Encoder Pretraining}
We use MoCo to pretrain our C3D+TCN encoder in an unsupervised manner. The original videos are input as support and videos augmented with ~\ref{augmentation} as the query. Then MoCo trains the encoder by a contrastive loss~\cite{moco} by comparing the query with multiple supports, both positive and negative. This enables the encoder to recognize and extract robust key features from the videos.

MoCo updates the encoder in a momentum way, so that the model is capable of building a consistent memory bank of recently seen data:
\begin{equation}
    \begin{aligned}
        \theta_k = m\theta_k + (1-m)\theta_q
    \end{aligned}
\end{equation}
where $m$ is the momentum coefficient, $\theta_k$ and $\theta_q$ are parameters of the key encoder and query encoder. During back propagation, only $\theta_q$ is updated, and then $\theta_k$ is updated using this momentum mechanism.

\subsection{Loose Action Alignment}\label{variable-length}

For a video $V$ with undefined length, we utilize sliding windows to segment the video $V$ to a set of windows of fixed-length $\{W_1,W_2,\cdots, W_n\}$. After the video embedding, each query window feature $\tilde{W^q_{k}}$ will be compared with a weighted aggregation of support class windows. Thus the classification probability of few shot class $\theta$ of $\tilde{W^q_{k}}$ will be represented as $P(\theta|W_k)$:
\begin{equation}
    \begin{aligned}
    P(\theta|W_k) &= g(\tilde{W^q_{k}}, A(\tilde{W^s_j},j\in C_{\theta}))\\
    \end{aligned}
\end{equation}
where $g(\cdot)$ is the window-wise relational convolution, $A(\cdot)$ is the attention over windows per class which will be detailed in Section~\ref{ap}.

After obtaining the class probability for each window $S_{k}^{\theta}$, two losses will be computed. The Connectionist temporal classification (CTC) loss is computed for each query video, by aligning all the windows of the query video sequentially, and take the negative likehood afterward. The standard MSE loss over one-hot label will also be computed by adding up the each window's probability for each class:
\vspace{-10pt}
\begin{equation}
    \begin{aligned}
        \mathcal{L}_{CTC} &= -P(V,l)
        = -\sum_{\Theta:\kappa(\Theta)=l}\prod_{t=1}^T P(\theta_t|W_t)\\
        \mathcal{L}_{MSE} &= \frac{1}{T}\sum_{t=1}^T\sum_{\theta \in \Theta}(\mathbbm{1}(\theta, l)-P(\theta|W_t))^2\\
        \mathcal{L} &= \mathcal{L}_{CTC} + \lambda * \mathcal{L}_{MSE}
    \end{aligned}
\end{equation}
where $l$ refers to the correct label of the video $V$, and $\kappa(\cdot)$ is the union of the window class trajectories leading to the right label.

The sliding windows trained with CTC loss can effectively solve the alignment issue for the videos in the temporal domain, which is robust against object occlusions, disocclusion, and other video instabilities. While the MSE loss can accelerate the training at the beginning. A weighted combination of the two loss provides optimized convergence of the model. At the test time, beam search decoding is performed to retrieve the video label. 

\subsection{Attention Pooling}
\label{ap}
Since multiple windows are spread across the support videos to extract support features, a pooling operation is necessary to generate fixed-size final representations for the relation network where they will be compared with the query. 

\begin{figure}[t] 
    \begin{center}
        \includegraphics[width=0.45\textwidth]{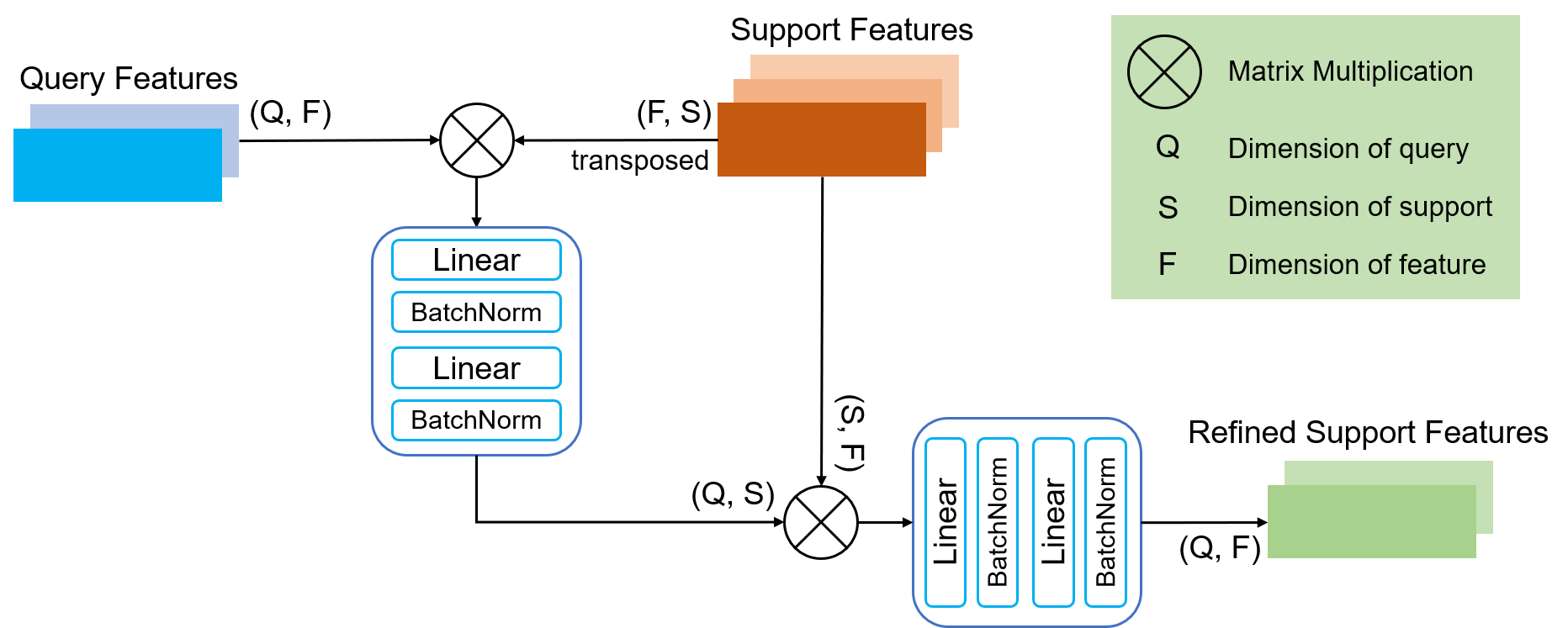}
        \hspace{0.09\textwidth}
        \includegraphics[width=0.45\textwidth]{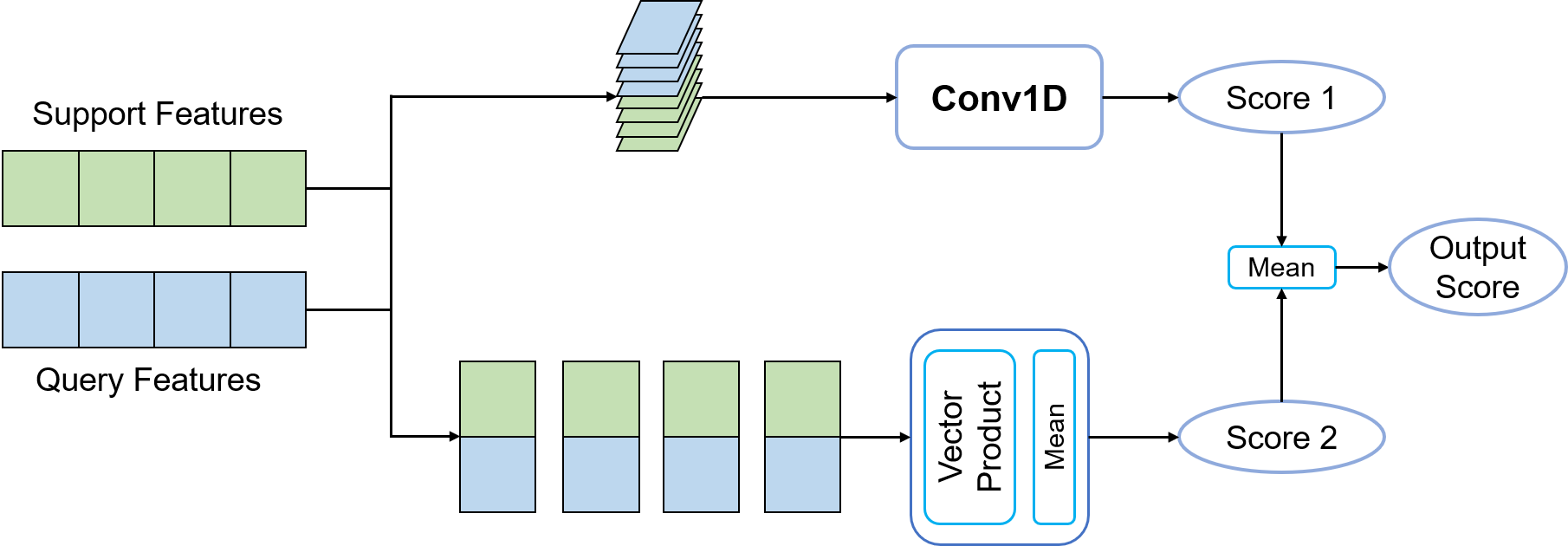}
        \begin{minipage}[t]{0.45\linewidth}\centering\subcaption{Attention Pooling}\label{fig:attention}\end{minipage}
        \hspace{0.09\textwidth}
        \begin{minipage}[t]{0.45\linewidth}\centering\subcaption{Multi-head Relation Network}\end{minipage}
    \end{center}
    \vspace{-7pt}
    \caption{Attention Pooling and Multi-head Relation Network}
\end{figure}

Commonly used methods are straightforward max-pooling, average-pooling and self-attention. In our work, we propose an attention pooling module. As illustrated in Figure~\ref{fig:attention}, our specially-designed attention pooling takes both the support feature $\mathcal{S} \in \mathbb{R}^{S\times F}$ and query feature $\mathcal{Q} \in \mathbb{R}^{Q\times F}$ as input and computes new support features $\mathcal{S}^\prime$ as follows: 
\vspace{2pt}
\begin{equation}
    \label{refine-support}
    \begin{aligned}
        \mathcal{S}^\prime &= f_2(f_1(\mathcal{Q}\cdot \mathcal{S}^T)\cdot \mathcal{S})\\
    \end{aligned}
\end{equation}
\vspace{2pt}
where $f_i(\cdot)$ are linear transformations. The idea is to introduce query features to the pooling procedure by multiplying query features with support features transpose, which will generate a weight matrix $\mathcal{W} \in \mathbb{R}^{Q\times S}$. Each entry $\mathcal{W}_{i,j}$ represents the weight of $j^{th}$ support window to the $i^{th}$ query window. Then the product of this weight matrix and original support features can be seen as the refined support features. Two linear functions $f_1$ and $f_2$ are added to provide some learnable parameters.

In addition, to support feature refinement, we propose to refine query features by support features. Specifically, the same Equation \ref{refine-support} can be applied except we swap the support and query. This mutual refinement strategy can enhance the prediction performance.

\subsection{Multi-Head Relation}
Since the few shot models cannot retain the action class features in the FC layers, as the unseen classes keep coming, it's crucial to compare the feature similarity of the query video. The previous relation network ~\cite{relation-net} used in few shot action recognition compares the video-wise similarity with a number of convolutional layers, whose performance significantly as the length of video increases. We extend this network to a multi-head relation network in two ways.

First, we reduce the convolution kernel size to 1 and introduce the FC layer in our multi-head relation since the extracted features no longer retain any spatial structure. Second, we add one more computation layer on top of the original Conv+FC layers, which is a window-wise vector product. This provides a more localized comparison between the support and query. The final output of multi-head relation network is the sum of the probabilities obtained from both methods.

\section{Experiments}
\subsection{Dataset}
We test our model on three datasets including Human-Centric Atomic Action (HAA), Finegym (Gym288), and Moments in Time (MIT). MIT is a general atomic action dataset with over one million videos from diverse classes~\cite{mit}. Finegym is a recent dataset which focuses on fine-grained gymnastic action classes~\cite{finegym}. HAA provides human-centric videos, with a high average of 69.7\% detectable joints ~\cite{haa}. These datasets are not constructed specifically for few-shot learning, so we reorganize them to suit our few-shot tasks.

\noindent\textbf{HAA \& Gym288}. Instead of splitting videos into training and test set, we split action classes into training and test set for our few-shot tasks. Consequently, we have 310/156 classes in our HAA training/test set, and 231/57 in our Gym288 respectively.

\noindent\textbf{MIT}. The total number of videos in the MIT dataset is huge so we build a mini-MIT for our experiment. Each action class in mini-MIT has 60 videos, half from the original training set, and another half from the original validation set. Like HAA and Gym288, our mini-MIT has 272/67 classes for training and test respectively.

\subsection{Model Performance}

\begin{table}[h]
    \begin{center}
    \begin{tabular}{cccc}
        \hline
        Model & \textbf{HAA}~\cite{haa} & \textbf{Gym288}~\cite{finegym} & \textbf{mini-MIT}~\cite{mit}\\
        \hline
        \hline
        Ours & \textbf{80.68} & \textbf{75.84} & \textbf{61.77}\\
        \hline
        \multirow{2}{*}{SOTA} & 55.33 / 75 & \multirow{2}{*}{73.7} & 31.16 / 57.67\\
        & top1 / top3 & & top1 / top3\\
        \hline
    \end{tabular}
    \end{center}
    \caption{Top-1 accuracy (\%) on three datasets under a 3-way 5-shot setting compared with state-of-the-art performance in~\cite{haa,finegym,mit} where their models are trained in full supervision.}
    \label{tab:exp}
\end{table}

Table~\ref{tab:exp} tabulates our results, showing that our few-shot model has a leading performance compared with the state-of-the-art methods on all the three datasets trained in full supervision. Note that our model is few-shot which has access to only a very limited amount of data.

Further, the HAA and Gym288 are the atomic human-centric datasets, while the former has similar background and the latter's background is heterogeneous. The MIT, on the other hand, doesn't follow a strict human-centric manner. The outperforming result over all 3 datasets presenting the potentials of our innovative loose action alignment and attention-based feature aggregation over a general understanding of atomic action. Besides, the multi-head relational network's improvement is not limited to human-centric datasets, showing the importance of local comparison on action similarity.

\subsection{Ablation Study}

\begin{table}[h]
    \begin{center}
    \begin{tabular}{cccc}
        \hline
        Model & \textbf{HAA} & \textbf{Gym288} & \textbf{mini-MIT}\\
        \hline
        \hline
        w/o pretraining & \textbf{81.84} & 72.42 & 42.48\\
        \hline
        w/o AP & 77.53 & 66.06 & 39.93\\
        \hline
        w/o RN & 63.22 & 69.30 & 37.37\\
        \hline
        \hline
        Ours & 80.68 & \textbf{75.84} & \textbf{61.77}\\
        \hline
    \end{tabular}
    \end{center}
    \caption{Performances without the key features in our model.}
    \label{tab:ablation}
\end{table}

Table \ref{tab:ablation} shows our ablation study result, and proves the effectiveness of our semi-supervised training, attention-based feature aggregation and multi-head video comparison.

Specifically, the unsupervised action encoder may experience relatively less accuracy drop in human-centric datasets such as HAA where the action features are better aligned.
However, on more general datasets, the human-centric augmentation in Section ~\ref{augmentation} shows greater importance and the ablation accuracy drops significantly on mini-MIT. Besides, the performance on HAA and Gym288 drops less compared with that on mini-MIT, indicating our model's better representativeness over a general set of action data.

\section{Conclusion}
This paper introduces a novel semi-supervised few-shot atomic action recognition model. 
We propose to use the sliding windows and CTC alignment to make the model more robust to coarse temporal annotation. Operating in the few-shot setting, our model can effectively alleviate human annotation effort. Moreover, 
we propose the attention pooling and multi-head relation module to achieve better feature refinement and comparison. By incorporating unsupervised and contrastive video embedding, our few-shot model produces state-of-the-art performance comparable to previous models trained in full supervision.

{\small
\bibliography{egbib}
}

\end{document}